\documentclass[letterpaper]{article}
\usepackage{uai2019}
\usepackage[margin=1in]{geometry}

\usepackage{times}
\usepackage[utf8]{inputenc} % allow utf-8 input
\usepackage[T1]{fontenc}    % use 8-bit T1 fonts
\usepackage{hyperref}       % hyperlinks
\usepackage{url}            % simple URL typesetting
\usepackage{booktabs}       % professional-quality tables
\usepackage{amsfonts,amsthm}       % blackboard math symbols
\usepackage{nicefrac}       % compact symbols for 1/2, etc.
\usepackage{microtype}      % microtypography
\usepackage{amsmath}
\usepackage{dirtytalk}
\usepackage{tikz}

\usepackage{amsmath}
\usetikzlibrary{bayesnet}
\usetikzlibrary{positioning}

\newcommand{\bpsi}{\boldsymbol{\psi}}

\newcommand{\bgamma}{\boldsymbol{\gamma}}
\newcommand{\bphi}{\boldsymbol{\phi}}
\newcommand{\bPsi}{\boldsymbol{\Psi}}
\newcommand{\blambda}{\boldsymbol{\lambda}}
\newcommand{\bOmega}{\boldsymbol{\Omega}}

\newcommand{\bt}{\boldsymbol{t}}
\newcommand{\by}{\boldsymbol{y}}
\newcommand{\bz}{\boldsymbol{z}}
\newcommand{\bw}{\boldsymbol{w}}

\newcommand{\params}{\boldsymbol{\theta}}

\newcommand{\ci}{\perp\!\!\!\perp}

\newcommand{\eqn}[1]{\begin{align}#1\end{align}}
\newcommand{\eq}[1]{\begin{align*}#1\end{align*}}

\renewcommand{\P}[1]{\mathbb{P}(#1)}

\definecolor{green}{HTML}{60ba46}
\definecolor{blue}{HTML}{1749b3}
\definecolor{orange}{HTML}{fa9d00}

\newcommand{\G}{\textcolor{blue}{G}}
\newcommand{\Gin}{\textcolor{green}{G_{\overline{T}}}}
\newcommand{\Gout}{\textcolor{orange}{G_{\underline{T}}}}

\title{Replacing the do-calculus with Bayes rule}

\author{ {\bf Finnian Lattimore\thanks{finn.lattimore@gmail.com}} \\
The Gradient Institute\\
Sydney \\
\And
{\bf David Rohde\thanks{d.rohde@criteo.com}}  \\
Criteo AI Lab \\
Paris \\
}

\begin{document}

\maketitle

\begin{abstract}
%It has been previously argued that an extension to probability known as the do-calculus is required in order to solve causal problems using probability theory. Contrary to this we demonstrate through four worked examples that equivalent causal estimation can be made using both the do-calculus and Bayesian statistics.  This takes an important step into unifying these two important approaches for causal inference.

The concept of causality has a controversial history. The question of
whether it is possible to represent and address causal problems with
probability theory, or if fundamentally new mathematics such as the do-calculus is required
has been hotly debated, e.g. \cite{pearl2001bayesianism} states \textit{`the building
  blocks of our scientific and everyday knowledge are elementary facts
  such as “mud does not cause rain” and “symptoms do not cause
  disease” and those facts, strangely enough, cannot be expressed in
  the vocabulary of probability calculus'}. This has lead to a
dichotomy between advocates of causal graphical modeling and the
do-calculus, and researchers applying Bayesian methods.  In this paper
we demonstrate that, while it is critical to explicitly model our
assumptions on the impact of intervening in a system, provided we do
so, estimating causal effects can be done entirely within the standard
Bayesian paradigm. The invariance assumptions underlying causal
graphical models can be encoded in ordinary Probabilistic graphical models,
allowing causal estimation with Bayesian statistics, equivalent to the
do-calculus. Elucidating the connections between these approaches is a
key step toward enabling the insights provided by each to be combined
to solve real problems.  
\end{abstract}

\section{Introduction}

The do-calculus \cite{pearl1995causal} is a powerful body of theory
that provides three additional rules for probability theory on the
basis that probability theory alone is not sufficient for solving
causal problems (an argument prosecuted forcefully by Pearl in
several places e.g. \cite{pearl2001bayesianism}).  In this paper we
provide side by side analysis of four classic causal problems: the two
cases responsible for Simpson's reversal, and two cases with
unobserved confounders.  These four analyses are strongly
suggestive that the do-calculus and Bayesian inference can both be
used in order to make causal estimates, although this is not to
suggest that each approach does not have its strengths and
weaknesses. 

A fully Bayesian approach leverages a vast body of existing
research and is able to account for finite sample uncertainties.  On
the other hand the do-calculus has simpler graphs and is sometimes a
more direct approach, also some of the results pertaining to
unobserved confounders were discovered using the do-calculus (in
particular results for front door adjustment and M-Bias) and while
we show these results can be transferred in the Bayesian paradigm the
mechanism for systematically doing so in a tractable manner remains unclear. 

While there is much to be discussed about the similarities and
differences between the two approaches we mostly leave this outside
scope and simply provide the four examples side by side.  The paper
has the following structure, in Section 2 we outline the two
methodologies in a sufficiently general framework that either could be
used to solve causal problems.  In Section 3 we outline the two fully
observed problems, here we focus on Simpson's paradox.  In Section 4
we outline two  problems involving unobserved confounders; the
causality non-identifiable case and the front door rule; concluding
remarks are made in Section 5. 

\section{Two schools of thought}
%Connecting Bayesian inference with causal graphical models

\subsection{Probabilistic graphical models}

Probabilistic graphical models (PGMs) combine graph theory with
probability theory in order to develop new algorithms and to present
models in an intuitive framework \cite{jordan2004graphical}. A Probabilistic graphical model is a directed acyclic graph over variables, which represents how the joint distribution over these variables may be factorized. In particular, any \emph{missing} edge in the graph must correspond to a conditional independence relation in the joint distribution. There are multiple valid Probabilistic graphical model representations for a given joint distribution. For example, any joint distribution over two variables $(X,Y)$ may be represented by both $X \rightarrow Y$ or $X \leftarrow Y$.

%hierarchical models are a similar framework \cite{gelman2013bayesian}. 

%Although PGMs may be used in many contexts we adopt them here in order to illustrate similarities in differences between the direct probability calculations for solving causal problems using Bayesian inference on a PGM with the use of the do-calculus on the CGM.  It is an important observation that the PGM operates on an extended space combining the mutilated and un mutilated graph into a single framework. 

\subsection{Causal Graphical Models And The Do-Calculus}

A causal graphical model (CGM) is a Probabilistic graphical model, with the additional assumption that a link $X \rightarrow Y$ means $X$ causes $Y$. Think of the data generating process for a CGM as sampling data first for the exogenous variables (those with no parents in the graph), and then in subsequent steps sampling values for the children of previously sampled nodes. An atomic intervention in such a system that sets the value of a specific variable $T$ to a fixed constant corresponds to removing all links into $T$ - as it is now set exogenously, rather than determined by its previous causes. It is assumed that everything else in the system remains unchanged,  in particular the functions or conditional distributions that determine the value of a variable given its parents in the graph. In this way, a CGM encodes more than the factorization (or conditional independence structure) of the joint distribution over its variables; It additionally specifies how the system responds to atomic interventions. 

A CGM describes how the structure of a system is modified by an intervention. However, answering causal queries such as \textit{"what would the distribution of cancer look like if we were able to prevent smoking?"} requires inference about the distributions of variables in the post-interventional system. The do-notation is a short-hand for describing the distribution of variables post-intervention and the do-calculus is a set of rules for identifying which (conditional) distributions are equivalent pre and post-intervention. If it is possible to derive an expression for the desired post-interventional distribution purely in terms of the joint distribution over the original system via the do-calculus then the causal query is identifiable, meaning assuming positive density and infinite data we obtain a point estimate for it. The do-calculus is complete; A query is identifiable if and only if it can be solved via the do-calculus \cite{shpitser2006identification,huang2006pearl}. 

Here we present the do-calculus in a simplified form that applies to interventions on single variables - which is sufficient for the examples presented in this paper. The full form of the do-calculus applies to interventions on any subset of variables - see \cite{pearl1995causal,pearl2009,peters2017elements}. 

\paragraph*{The do-calculus} Let $G$ be a CGM, $\Gin$ represent $G$ post-intervention (i.e with all links into $T$ removed) and $G_{\underline{T}}$ represent $G$ with all links \emph{out of} $T$ removed. Let $do(t)$ represent intervening to set a single variable $T$ to $t$,

\subparagraph{Rule 1:} $\P{y|do(t),z,w}=\P{y|do(t),z}$ if $Y\ci W|(Z,T)$ in $\Gin$
\subparagraph{Rule 2:} $\P{y|do(t),z}=\P{y|t,z}$ if $Y\ci T|Z$ in $\Gout$
\subparagraph{Rule 3:} $\P{y|do(t),z}=\P{y|z}$ if $Y\ci T|Z$ in $\Gin$, and $Z$ is not a decedent of $T$.

\subsection{Representing a Causal Problem with a Probabilistic graphical model}

While PGMs and CGMs may appear similar, there are key differences
between them, both in the information they represent and how they are
typically applied. CGMs are used to determine if a given query is
identifiable and to obtain an expression for it in terms of the
original joint distribution - with estimation of this expression a
follow up step; latent variables are introduced to capture dependence
induced by unobserved variables that may complicate identification of
causal effects and links are not reversible. By contrast in PGMs links
can be reversed, model specific details for estimation - including
plates \& parameters - are included graphically, and latent variables
(of a specific form) are introduced for computational reasons, usually
to coerce the model into complete data exponential family form. 

To represent an intervention with an ordinary Probabilistic graphical model, we must explicitly model the pre and post intervention systems and the relationship between them.   %, see figure \ref{fig:bayes_case1} for an example.
\paragraph*{Algorithm 1: CausalBayesConstruct}\label{Alg:causebayesconstruct}~\\ \emph{Input}: Causal graph $G$ and intervention $do(T=t)$. ~\\\emph{Output}: Probabilistic graphical model representing this intervention
\begin{enumerate}
\item Draw the original causal graph $G$ inside a plate indexed from $1, ... M$ to represent the data generating process.
\item For each variable $V\in G$, parameterize $P(V|parents(V))$ by adding a parameter $\theta_V$ with a link into $V$. 
\item Draw the graph after the intervention by setting $T=t$ and
  removing all links into it. %Make any nodes that are ancestors of
                              %$T$ latent, since they cannot be
                              %observed before selecting $t$. % What
                              %does this mean???
Rename each of the variables to distinguish them from the variables in the original graph, e.g. $X$ becomes $X^*$.
\item Connect the two graphs linking $\theta_V$ to the corresponding variable $V^*$ in the post-interventional graph, for each $V$ excluding $T$.

%with a shared parameter representing
%  $P(V|{\rm parents}(V))$ for each pair of variables ($V$,$V^*$) except
%  $T$,  The conditional distribution that determines the value of each %variable given its parents is assumed to be unchanged for all other variables.

\end{enumerate}

A PGM constructed with Algorithm 1 represents exactly the same assumptions about a specific intervention as the corresponding CGM, see Figures~\ref{fig:cgm1} and \ref{fig:pgm1} for an example. We have just explicitly created a joint model over the system pre and post-intervention, which allows the direct application of standard statistical inference, rather than requiring additional notation and operations that map from one to the other - as the do-calculus does. The Bayesian model is specified by the parameterization of the conditional distribution of variables given their parents, and priors may be placed on the parameters $\params$. The fact that the parameters are shared for all pairs of variables $(V,V^*)$ excluding $T$, captures the assumption that all that is changed by the intervention is the way $T$ takes its value - the conditional distributions for all other variables given their parents are invariant. 

Despite its simplicity we are unaware of a direct statement of Algorithm 1, it is related to twin networks \cite{pearl2009} and augmented directed acyclic graphs \cite{dawid2015statistical} but is distinct from both.

\subsection{Causal Inference with Probabilistic graphical models}
The result of Algorithm 1 is a Probabilistic graphical model on which we can do inference with standard probability theory rather than the do-calculus, and which has properties such as arrow reversal (by the use of Bayes rule). To infer causal effects we compute a predictive distribution for the quantity of interest in the post-intervention graph using Bayes rule, integrating out all parameters, latent variables and any observed variables that are not of interest, for each setting of the treatment $T=t^*$. 

%For each setting of the treatment $T=t^*$ compute a predictive distribution for the quantity of interest in the post intervention graph using Bayes rule, integrating out all parameters, latent variables and any observed variables that are not of interest.  %Conditional average treatment effects (or
%personalized treatment effects will condition on covariates where
%average treatment effects will integrate them out.

\newcommand{\Dpost}{\boldsymbol{v^*}}
\newcommand{\Dpre}{\boldsymbol{v}}

To make this procedure clearer, let $\boldsymbol{V}$ be the set of variables in the original causal graph $\G$, excluding the variable we intervene on, $T$, and $\boldsymbol{V^*}$ be the corresponding variables in the post-interventional graph. We have:
\begin{itemize}
\item
$\params$: the set of model parameters.
\item
$\Dpre$: a matrix of the $M$ observations of variables $\boldsymbol{V}$, $(\boldsymbol{v}_1,...,\boldsymbol{v}_M)$ collected pre-intervention.
\item
$\bt$: a vector of the $M$ observed values of the treatment variable $T$, $t_1,..t_M$, and
\item
$\Dpost$: The variables of the system post-intervention.
\item
$t^*$: the value that the intervened on variable $T$ is set to. 
\item 
$Y^*\in {\Dpost}$: the variable of interest post-intervention.

%as we set the value of $t^*$ it does not have a distribution in fact we usually want to optimize $t^*$ in order to produce a desired outcome on $y^*$.
\end{itemize}

The goal is to infer the value of the unobserved post-interventional
distribution over $\Dpost$, given the observed data and $(\Dpre,\bt)$
and a selected treatment $t^*$. By construction, conditional on the
parameters $\params$, the post-interventional variables $\Dpost$ are
independent of data collected pre-intervention $(\Dpre,\bt)$. The
value of the intervention $t^*$ is set exogenously\footnote{Also $t^*$
  has no marginal distribution - it is a constant set by the
  intervention} - so is independent of both $\params$ and
$(\Dpre,\bt)$. This ensures joint distribution over
$(\Dpre,\bt,\Dpost,\params)$ factorize into three terms: a prior over
the parameters $\P{\params}$, the likelihood for the original system
$\P{\Dpre,\bt|\params}$, and a predictive distribution for the
post-interventional variables given parameters and intervention
$\P{\Dpost|\params,t^*}$:

\eq{
\P{\Dpre,\bt,\Dpost,\params|t^*} = \P{\params}\P{\Dpre,\bt|\params}\P{\Dpost|\params,t^*}
}

%The first term, $\P{\params}$ is the prior over the model parameters, the second is the likelihood, and the third

We then marginalize out $\params$,
\eqn{
\P{\Dpre,\bt,\Dpost|t^*} = \int_{\params}\P{\params}\P{\Dpre,\bt|\params}\P{\Dpost|\params,t^*}d\params
}
and condition on the observed data $(\Dpre,\bt)$,
\eqn{
\P{\Dpost|\Dpre,\bt,t^*} &= \frac{\P{\Dpre,\bt,\Dpost|t^*}}{\P{\Dpre,\bt|t^*}} \nonumber \\
&= \int_{\params}\frac{\P{\params}\P{\Dpre,\bt|\params}}
{\P{\Dpre,\bt}}\P{\Dpost|\params,t^*}d\params \nonumber \\
&= \int_{\params}\P{\params|\Dpre,\bt}\P{\Dpost|\params,t^*}d\params.
\label{eqn:post_given_observed_general}
}
Finally, if the goal is to infer mean treatment effects\footnote{We could also compute conditional treatment effects by first conditioning on selected variables in $\Dpost$.} on a specific variable post-intervention $Y^*$, we can marginalize out the remaining variables in $\boldsymbol{V^*}$, 
\eqn{
\P{Y^*|\Dpre,\bt,t^*}
&= \int_{\params}\P{\params|\Dpre,\bt}\sum_{\boldsymbol{V^*}\backslash Y^*}\P{\Dpost|\params,t^*}d\params.
\label{eqn:post_given_observed_general}
}

If there are no latent variables in $\G$, assuming positive density over the domain of $(\Dpre,\bt)$ and a well defined prior $\P{\params}$, the likelihood  $\P{\Dpre,\bt|\params}$ will dominate, and the posterior over the parameters $\P{\params|\Dpre,\bt}$ will become independent of the prior at the infinite data limit. The term $\P{\Dpost|\params,t^*}$ can be expanded into a product of terms of the form $\P{V^*|parents(V^*),\params}$ following the factorization implied by the post-interventional graph. From step (3) of Algorithm 1 each of these terms are equal to the corresponding terms $\P{V|parents(V),\params}$, giving results equivalent to Pearl's truncated product formula \cite{pearl2009}.

% It looks a little different because in Pearl's version, V includes T. It becomes notationally messy to explicitly show equivalence ...

The presence of latent variables in $\G$ adds complications which we defer to Section 4.

%Marginalizing out $z^*$ to obtain the average treatment effect gives:
%
%\eqn{
%\P{y^*|\by,\bz,\bt,t^*} = \int_{\theta}\P{\theta|\by,\bz,\bt}\sum_{z^*}\P{y^*,z^*|t^*,\theta}d\theta
%\label{eqn:general_bayesian}
%}

%By construction, the data generating process is, conditional on the
%parameters, independent of the intervention variables; implying the
%above factorization into three parts.  The first part is a  component predicting the
%outcome for a counterfactual treatment ($t^*$), the second the likelihood
%and the third the prior on the parameters. % FIXME Finn is
                                % this ok now???  !!!!!!!!!!!!!!!!!!!!!!!!!!!!
%Moreover, usually $\P{\by,\bz,\bt|\theta}$ will further factor in a way specified
%by the structure of the graph, providing further problem specific simplifications.
%This formulation covers the three node fully observed case (multiple
%nodes simply involve $Z$ and $\theta$ being split into multiple parts).  Unobserved
%confounders / latent variables are a further complication and are
%deferred to Section 4.  It is sometimes relevant and interesting to
%produce a conditional average treatment effect which conditions rather
%than averages over $z^*$, while our framework covers this case we do
%not further consider it.

\section{Simpson's Paradox (Fully Observed)}
Simpson's paradox provides an excellent case study for demonstrating
that raw data cannot be used for inferring causality without further
assumptions. In this section, we show how we can infer treatment effects and resolve the paradox, with either the do-calculus or via Bayesian inference, and that these approaches yield equivalent results. Assume we have a table of data on some outcome $Y$ for two
different treatments ($T$) broken down by a third variable $Z$ as shown in Table \ref{tbl:example_data}.  

\begin{table}[hb!] %TODO can we force this table to be here? Otherwise we have to reference it.
\center
\begin{tabular}{ l l l r }
\toprule
$z$ & $t$ & $y$ & $N$ \\
\midrule \midrule
0 & 0 & 0 & 150\\
0 & 0 & 1 & 50\\
0 & 1 & 0 & 180\\
0 & 1 & 1 & 180\\
1 & 0 & 0 & 50\\
1 & 0 & 1 & 200\\ 
1 & 1 & 0 & 4\\
1 & 1 & 1 & 36\\
\bottomrule
\end{tabular}
\caption{Example Data}
\label{tbl:example_data}
\end{table}

% https://towardsdatascience.com/simpsons-paradox-how-to-prove-two-opposite-arguments-using-one-dataset-1c9c917f5ff9
% aa <-data.table(Z=c(0,0,0,0,1,1,1,1),T=c(0,0,1,1,0,0,1,1),Y=c(0,1,0,1,0,1,0,1),N=c(150,50,180,180,50,200,4,36))
% aa[,sum(N),list(T,Y)]

By estimating probabilities as past frequencies we obtain the following conditional probabilities: %and by both including and excluding the covariate $Z$ we
\eqn{
P(Y=1|Z=0,T=0) = 0.25 \nonumber\\
P(Y=1|Z=0,T=1) = 0.5 \nonumber\\
P(Y=1|Z=1,T=0) = 0.8\nonumber\\
P(Y=1|Z=1,T=1) = 0.9\nonumber 
}
\begin{tabular}{r r}
$P(Y=1|T=0) = 0.56$ & $P(Y=1|T=1) = 0.54$ \\
\end{tabular}

\paragraph*{The paradox} Treatment $T=0$ seems best overall, but if we
break the data down by $Z$ then, regardless of which value of $Z$ a patient has, treatment $T=1$ seems better. If we had to select a single treatment for everyone - which should it be? The key to resolving this question is to realize that what we care about in this setting is the expected outcome of \emph{intervening} in the system to set $T$, (in Pearl's notation $\P{Y|do(T)}$) rather than either of the conditional distributions $\P{Y|T}$ or $\P{Y|T,Z}$. As a result, which treatment is preferred hinges on causal assumptions, which we may specify using a CGM or a PGM.

\subsection{Simpson's Paradox Case 1}

\begin{figure}
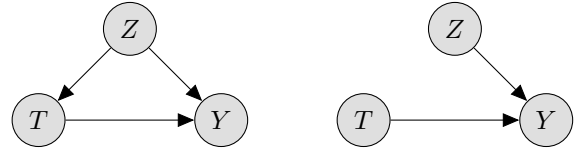

  \centering
  \tikz{ %
    \node[obs] (Zm) {$Z$} ; %
    \node[obs, below left=of Zm] (Xm) {$T$} ; %
    \node[obs, below right=of Zm] (Ym) {$Y$} ; %
    \edge {Zm} {Xm} ; %
    \edge {Zm} {Ym} ; %
    \edge {Xm} {Ym} ; %
  } \hspace{1cm}
  \tikz{ %
    \node[obs] (Zm) {$Z$} ; %
    \node[obs, below left=of Zm] (Xm) {$T$} ; %
    \node[obs, below right=of Zm] (Ym) {$Y$} ; %
    \edge {Zm} {Ym} ; %
    \edge {Xm} {Ym} ; %
  }
  \caption{A CGM of Case 1: Left observational, Right: mutilated}
  \label{fig:cgm1}  
\end{figure}

\begin{figure}[hb!]
\caption{The graphs $\G$, $\Gin$ and $\Gout$ for case 1.}
\label{fig:differentG}
\centering
\includegraphics[width=.4\textwidth]{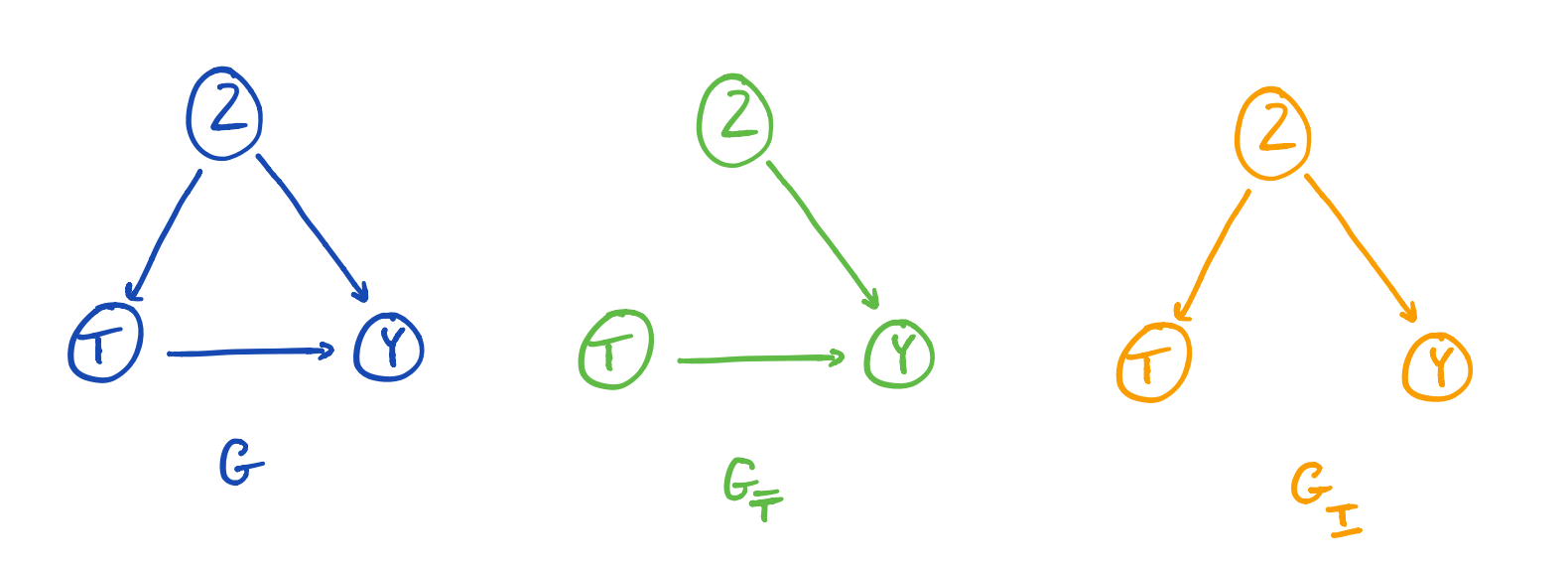}
\end{figure}

\begin{figure}
  \centering
  \tikz{ %
    \node[obs] (Zm) {$Z_m$} ; %
    \node[obs, below left=of Zm] (Xm) {$T_m$} ; %
    \node[obs, below right=of Zm] (Ym) {$Y_m$} ; %
    \edge {Zm} {Xm} ; %
    \edge {Zm} {Ym} ; %
    \edge {Xm} {Ym} ; %
    \plate[inner sep=0.25cm, xshift=-0.12cm, yshift=0.12cm] {plate1} {(Xm) (Zm) (Ym)} {m=1..M};
    \node[latent, below=of Ym] (Zn) {$Z^*$} ; %
    \node[obs, below left=of Zn] (Xn) {$T^*$} ; %
    \node[latent, below right=of Zn] (Yn) {$Y^*$} ; %
    \edge {Zn} {Yn} ; %
    \edge {Xn} {Yn} ; %
%    \plate[inner sep=0.25cm, xshift=-0.12cm, yshift=0.12cm] {plate1} {(Xn) (Zn) (Yn)} {n=M+1..N};
    \node[latent, right=of Ym] (gamma) {$\gamma$};
    \node[latent, left=of Xm] (phi) {$\phi$};
    \node[latent, below=of Xm] (theta) {$\psi$};
    \edge {gamma} {Zm,Zn};
    \edge {phi} {Xm};
    \edge {theta} {Ym,Yn};    
  }
  \caption{A PGM of Case 1}
  \label{fig:pgm1}  
\end{figure}

Imagine that our observations are generated by CGM given in Figure~\ref{fig:cgm1}, where the covariate $Z$ is a cause of both $T$ and $Y$.  Applying the rules of the do-calculus gives:
\eqn{
\P{&y|do(T=t)} = \sum_z \P{y,z|do(T=t)} \nonumber \\
&= \sum_z \P{y|do(T=t),z}\P{z|do(T=t)}\nonumber \\
&= \sum_z \underbrace{\P{y|t,z}}_{\text{Rule  2:($Y \ci T|Z$ in $\Gout$) }}\underbrace{\P{z}}_{\text{Rule 3:($Z \ci T$ in $\Gin$)}}\nonumber
}
See Figure~\ref{fig:differentG} for $\G$, $\Gin$ and $\Gout$ in case 1. %Then parameterizing $\P{y|t,z}$ with $\phi$ and $\P{z}$ with $\gamma$ and estimating:

To find the same solution using a Bayesian approach we first
apply \textbf{CausalBayesConstruct} on CGM Figure~\ref{fig:cgm1} to
produce the PGM in Figure~\ref{fig:pgm1}.  We then explicitly parameterize the
model and  write out the three model components, the post intervention predictive, the
likelihood component and the prior component.  

We use the following parameterization:
\eqn{
&\P{y|t,z,\bpsi} =\bpsi_{t,z}^{y}(1-\bpsi_{t,z})^{1-y} = \P{y^*|t^*,z^*,\bpsi}, \nonumber\\
&\P{z|\bgamma} =\bgamma^{z}(1-\bgamma)^{1-z} = \P{z^*|\bgamma},\nonumber\\
&\P{t|z,\bphi} =\bphi_{z}^{t} (1-\bphi_z)^{1-t}. \nonumber
}   % DR fix here!
The post intervention predictive is:
\eqn{
\P{y^*|t^*,\bpsi,\bgamma} = & \sum_{z^*}\P{y^*|t^*,z^*,\bpsi}\P{z^*|\bgamma}. \nonumber \\
=&\sum_{z} \bpsi_{t^*,z}^{y^*}(1-\bpsi_{t^*,z})^{1-y^*} \bgamma^z(1-\bgamma)^{1-z}\nonumber
}
The likelihood component is:
\eqn{
\P{&\by|\bt,\bz,\bpsi}\P{ \bt|\bz,\bphi }\P{\bz|\bgamma} \nonumber\\
=&\prod_m \bpsi_{t_m,z_m}^{y_m}(1-\bpsi_{t_m,z_m})^{1-y_m}\nonumber\\
& \times \bphi_{z_m}^{t_m} (1-\bphi)^{1-z_m}\bgamma^{z_m}(1-\bgamma)^{1-z_m}
\nonumber 
}

By de Finetti's strong law of large numbers \cite{de1980foresight} as $M\rightarrow \infty$ the posterior concentrate
on a single point $\hat{\bpsi},\hat{\bphi},\hat{\bgamma}$ and
${\hat{\bpsi}_{t,z}^{y}(1-\hat{\bpsi}_{t,z})^{1-y} \rightarrow
\P{y|t,z}}$ and 
${\hat{\bgamma}^z(1-\hat{\bgamma})^{1-z} \rightarrow \P{z}}$, consequently:
\eqn{
&{\rm lim}_{M\rightarrow \infty} \sum_z \hat{\bpsi}_{t^*,z}^{y^*}(1-\hat{\bpsi}_{t^*,z^*})^{1-y^*}
\hat{\bgamma}^z(1-\hat{\bgamma})^{1-z} \nonumber\\
&\rightarrow \sum_z \P{y|t,z}\P{z}. \nonumber
}
Which demonstrates the agreement between the Bayesian solution and the solution
found using the do-calculus at large samples.

This convergence is usually very fast and good agreement will also be
found for low sample sizes (where instead of using the point estimate
we integrate over the posterior) e.g. under uniform priors the posterior of
the parameters will have Beta distributions; and the predictive
distribution giving the causal inference can be computed using the
``Laplace smoothing algorithm'' which involves adding one to the
counts before normalizing.

Returning to the numerical example applying the do-calculus using the
maximum likelihood algorithm we obtain:
\eqn{
\P{&y|do(T=t)}  \nonumber\\
&= \P{y|t,Z=0}\P{Z=0} +  \P{y|t,Z=1}\P{Z=1}\nonumber 
}
so
\eqn{
\P{Y=1|do(T=0)}  %&= \P{y|0,Z=0} \P{Z=0} +  \P{y|0,Z=1}\P{Z=1} \nonumber \\% delete this line
                       &=  \frac{50}{200} \frac{560}{850}
                       +\frac{200}{250} \frac{290}{850} \nonumber \\
& \approx 0.4376471 \nonumber 
} % 50/200*560/850 + 200/250*290/850

\eqn{
\P{Y=1|do(T=1)} % &= \P{y|1,Z=0}\P{Z=0} +  \P{y|1,Z=1}\P{Z=1}\nonumber \\ % delete this line
                       &=  \frac{180}{360} \frac{560}{850} +
                       \frac{36}{40} \frac{290}{850} \nonumber \\
& \approx 0.6364706 \nonumber 
}% 180/360*560/850 + 36/40*290/850
Assuming uniform priors and applying the Bayesian solution we obtain:
\eqn{
%Fix t^*=0  &= \P{y|0,Z=0} \P{Z=0} +  \P{y|0,Z=1}\P{Z=1} \nonumber \\% delete this line
\P{Y^*=1|T^*=0,\bt,\by,\bz} &=  \frac{51}{202} \frac{561}{852}
+\frac{201}{252} \frac{291}{852} \nonumber \\
& \approx 0.4386687 \nonumber 
} % 51/202*561/852 + 201/252*291/852
\eqn{
%Fix t^*=0 &= \P{y|1,Z=0}\P{Z=0} +  \P{y|1,Z=1}\P{Z=1}\nonumber \\ % delete this line
\P{Y^*=1|T^*=1,\bt,\by,\bz}&=  \frac{181}{362} \frac{561}{852}
+\frac{37}{42} \frac{291}{852} \nonumber \\
& \approx 0.630114 \nonumber 
}% 181/362*561/852 + 37/42*291/852

We see good agreement between the two methods with the only difference
being the prior impact due to the finite sample.  We also see that
$T=1$ is the better treatment (assuming that $Y=1$ is the desired
outcome). 

\section{Simpson's Paradox Case 2}

\begin{figure}
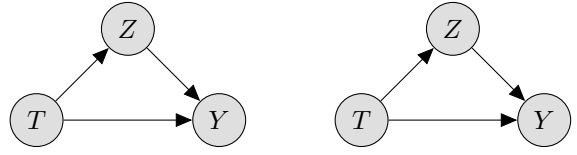

  \centering
  \tikz{ %
    \node[obs] (Zm) {$Z$} ; %
    \node[obs, below left=of Zm] (Xm) {$T$} ; %
    \node[obs, below right=of Zm] (Ym) {$Y$} ; %
    \edge {Xm} {Zm} ; %
    \edge {Zm} {Ym} ; %
    \edge {Xm} {Ym} ; %
  } \hspace{1cm}
  \tikz{ %
    \node[obs] (Zm) {$Z$} ; %
    \node[obs, below left=of Zm] (Xm) {$T$} ; %
    \node[obs, below right=of Zm] (Ym) {$Y$} ; %
    \edge {Xm} {Zm} ; %
    \edge {Zm} {Ym} ; %
    \edge {Xm} {Ym} ; %
  }

  \caption{A CGM of Case 2: Left observational, Right: mutilated}
  \label{cgm2}  
\end{figure}

\begin{figure}
%\caption{The graphs \textcolor{blue}{$G$}, \textcolor{green}{$G_{\overline{T}}$} and \textcolor{orange}{$G_{\underline{T}}$} for case 2.}
\caption{The graphs $\G$, $\Gin$ and $\Gout$ for case 2.}
\label{fig:differentG2}
\centering
\includegraphics[width=.4\textwidth]{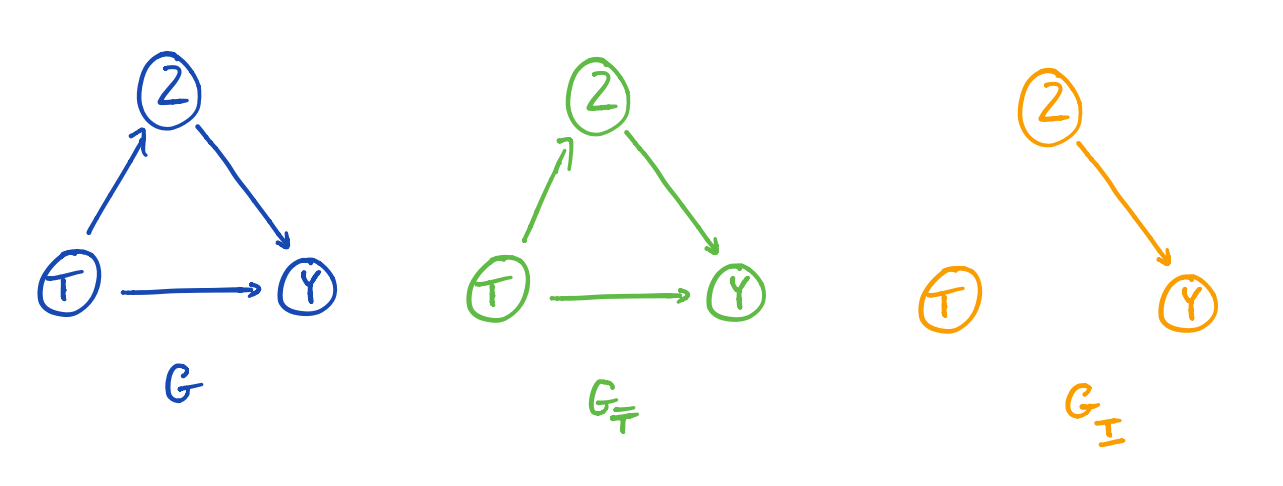}
\end{figure}

Imagine that our observations are generated by CGM given in
Figure~\ref{cgm2}.  Using the do-calculus we get the result in one
step: 
\eqn{
&\underbrace{\P{y|do(T=t)}  = \P{y|T=t}}_{\text{Rule 2}} \text{,
  since } Y \ci T \text{ in } \Gout   \nonumber \\
& = \sum_z\P{y,z|t}\nonumber
}
see Figure~\ref{fig:differentG2} to see the meaning of $\Gout$.

\begin{figure}
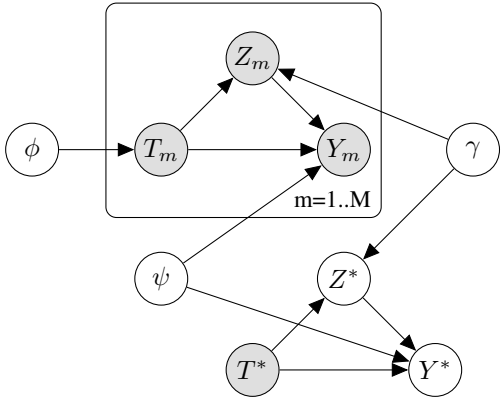

  \centering
  \tikz{ %
    \node[obs] (Zm) {$Z_m$} ; %
    \node[obs, below left=of Zm] (Xm) {$T_m$} ; %
    \node[obs, below right=of Zm] (Ym) {$Y_m$} ; %
    \edge {Xm} {Zm} ; %
    \edge {Zm} {Ym} ; %
    \edge {Xm} {Ym} ; %
    \plate[inner sep=0.25cm, xshift=-0.12cm, yshift=0.12cm] {plate1} {(Xm) (Zm) (Ym)} {m=1..M};
    \node[latent, below=of Ym] (Zn) {$Z^*$} ; %
    \node[obs, below left=of Zn] (Xn) {$T^*$} ; %
    \node[latent, below right=of Zn] (Yn) {$Y^*$} ; %
    \edge {Xn} {Zn} ; %
    \edge {Zn} {Yn} ; %
    \edge {Xn} {Yn} ; %
%    \plate[inner sep=0.25cm, xshift=-0.12cm, yshift=0.12cm] {plate1} {(Xn) (Zn) (Yn)} {n=M+1..N};
    \node[latent, right=of Ym] (gamma) {$\gamma$};
    \node[latent, left=of Xm] (phi) {$\phi$};
    \node[latent, below=of Xm] (theta) {$\psi$};
    \edge {gamma} {Zm,Zn};
    \edge {phi} {Xm};
    \edge {theta} {Ym,Yn};    
  }
  \caption{A PGM of Case 2}
  \label{pgm2}  
\end{figure}

To find the same solution using a Bayesian approach we first
apply \textbf{CausalBayesConstruct} on CGM Figure~\ref{cgm2} to
produce the PGM in Figure~\ref{pgm2}.  We then explicitly parameterize the
model and  write out the three model components, the post intervention predictive, the
likelihood component and the prior component.  

We use the following parameterization:
\eqn{
&\P{y|t,z,\bpsi} =\bpsi_{t,z}^{y}(1-\bpsi_{t,z})^{1-y} = \P{y^*|t^*,z^*,\bpsi},\nonumber\\
&\P{z|t,\bgamma}   = \bgamma_{t}^{z}(1-\bgamma_{t})^{1-z} = \P{z^*|t^*,\bgamma},\nonumber\\
&\P{t|\bphi}   = \bphi^{t}(1-\bphi)^{1-t}. \nonumber
}
The post intervention predictive is:
\eqn{
\P{&y^*|t^*,\bpsi,\bgamma} =  \nonumber \\
& = \sum_{z^*}\P{y^*|t^*,z^*,\bpsi}\P{z^*|t^*,\bgamma}\nonumber \\
& = \sum_z \bpsi_{t,z}^{y}(1-\bpsi_{t,z})^{1-y} \bgamma_{t}^z(1-\bgamma_{t})^{1-z}\nonumber
}
The likelihood component is:
\eqn{
\P{&\by|\bt,\bz,\bpsi}\P{ \bz|\bt,\bgamma }\P{ \bt| \bphi}=\nonumber \\
&\prod_m \bpsi_{t_m,z_m}^{y_m}(1-\bpsi_{t_m,z_m})^{1-y_m}\nonumber \\
& \times \bgamma_{t_m}^{z_m}(1-\bgamma_{t_m})^{1-z_m} \bphi^{t_m}(1-\bphi)^{1-t_m}\nonumber
}
Again by de Finetti's strong law of large numbers as
$M\rightarrow \infty$ the the posterior concentrate 
on a single point $\hat{\bpsi},\hat{\bphi},\hat{\bgamma}$ and
${\hat{\bpsi}_{t,z}^{y}(1-\hat{\bpsi}_{t,z})^{1-y} \rightarrow
  \P{y|t,z}}$ and
${\hat{\bgamma}_t^z(1-\hat{\bgamma}_t)^{1-z} \rightarrow \P{z|t}}$, consequently: 
\eqn{
&{\rm lim}_{M\rightarrow \infty} \sum_z \hat{\bpsi}_{t^*,z}^{y^*}(1-\hat{\bpsi}_{t^*,z})^{1-y^*}\hat{\bgamma}_{t^*}^z(1-\hat{\bgamma}_{t^*})^{1-z}\nonumber \\
&\rightarrow \sum_z \P{y,z|t} \nonumber
}
Showing that again there is large sample agreement between the two
methods, Similarly, convergence is usually very fast and there is
close agreement for even small samples.

Returning to the numerical example applying the do-calculus using the
maximum likelihood algorithm we obtain:
\eqn{
\P{Y=1|do(T=0)} = \frac{250}{450} \approx 0.5555556 \nonumber %\frac{50}{200} \frac{200}{450}  + \frac{200}{250} \frac{250}{450} %50/200*200/450 + 200/250*250/450
}
\eqn{
\P{Y=1|do(T=1)} & = \P{y|t} = \frac{216}{400} \nonumber \\ & \approx 0.54 \nonumber 
%= & \frac{180}{360} \frac{360}{400}  + \frac{36}{40} \frac{40}{400} % 180/360*360/400+36/40*40/400
}
Assuming uniform priors and applying the Bayesian solution again using
the Laplace smoothing result we obtain:
\eqn{
\P{Y^*=1|T^*=0,\bt,\by,\bz} %= %& P(Y=1|T=0,Z=0)P(Z=0|T=0) +P(Y=1|T=0,Z=1)P(Z=1|T=0) \\
&=  \frac{51}{202} \frac{201}{452}  + \frac{201}{252} \frac{251}{452}\nonumber \\
& \approx 0.5551989 \nonumber
}
\eqn{
\P{Y^*=1|T^*=1,\bt,\by,\bz} %= & P(Y=1|T=1,Z=0)P(Z=0|T=1) + P(Y=1|T=1,Z=1)P(Z=1|T=1)
& =  \frac{181}{362} \frac{361}{402}  + \frac{37}{42} \frac{41}{402} \nonumber \\
& \approx 0.5388534  \nonumber% 181/362*361/402+37/42*41/402
}
Again we see good agreement between the two methods with the only difference
being the prior impact due to the finite sample.  We also see that
$T=0$ is the better treatment.

Note that the distribution over $\P{z,t,y}$ is identical for both Case 1 and Case 2, and yet the optimal treatment differs. The paradox is resolved by understanding that the difference is due to different \emph{model assumptions} about the impact of intervening on $T$, and we have demonstrated that these assumptions can be expressed either with a CGM or an extended PGM.

\section{With Unobserved Confounders}

Unobserved confounders (or latent variables) are hidden variables that
can complicate causal inference at best and at worst render it
impossible.  While a direct attack using the pre-specified methodology
does allow Bayesian inference to solve these problems, this is achieved
by marginalizing out a complex latent variable the size of which grows
with the data set.  Usually the inclusion of the latent variable is
not viable and the model must be marginalize to remove it and
re-parameterized. Whether this is possible in a way that allows causation to
be identified depends on the structure of the graph.  If it isn't
possible to identify all parameters that have a causal impact then
prior distributions will have an impact even in the large data limit.

\subsection{When Causality Cannot Be Identified}

\begin{figure}
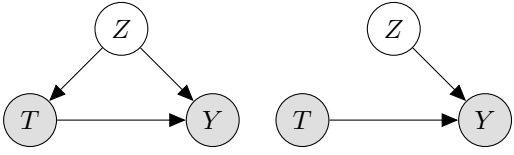

  \centering
  \tikz{ %
    \node[latent] (Zm) {$Z$} ; %
    \node[obs, below left=of Zm] (Xm) {$T$} ; %
    \node[obs, below right=of Zm] (Ym) {$Y$} ; %
    \edge {Zm} {Xm} ; %
    \edge {Zm} {Ym} ; %
    \edge {Xm} {Ym} ; %
  } \hspace{0.3cm}
  \tikz{ %
    \node[latent] (Zm) {$Z$} ; %
    \node[obs, below left=of Zm] (Xm) {$T$} ; %
    \node[obs, below right=of Zm] (Ym) {$Y$} ; %
    \edge {Zm} {Ym} ; %
    \edge {Xm} {Ym} ; %
  } 
  \caption{CGM Where Causality Is ``not identifiable'':  Left observational, Right: mutilated}
  \label{causation_hard}  
\end{figure}

\begin{figure}
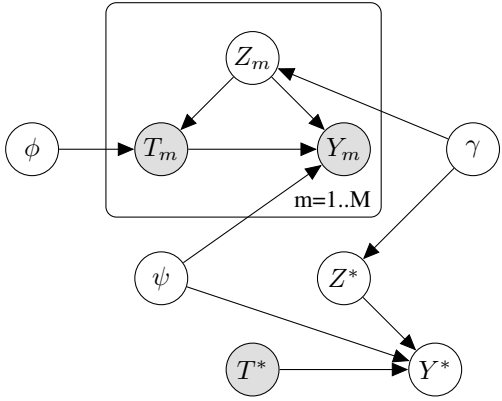

  \centering
  \tikz{ %
    \node[latent] (Zm) {$Z_m$} ; %
    \node[obs, below left=of Zm] (Xm) {$T_m$} ; %
    \node[obs, below right=of Zm] (Ym) {$Y_m$} ; %
    \edge {Zm} {Xm} ; %
    \edge {Zm} {Ym} ; %
    \edge {Xm} {Ym} ; %
    \plate[inner sep=0.25cm, xshift=-0.12cm, yshift=0.12cm] {plate1} {(Xm) (Zm) (Ym)} {m=1..M};
    \node[latent, below=of Ym] (Zn) {$Z^*$} ; %
    \node[obs, below left=of Zn] (Xn) {$T^*$} ; %
    \node[latent, below right=of Zn] (Yn) {$Y^*$} ; %
     \edge {Zn} {Yn} ; %
    \edge {Xn} {Yn} ; %
    \node[latent, right=of Ym] (gamma) {$\gamma$};
    \node[latent, left=of Xm] (phi) {$\phi$};
    \node[latent, below=of Xm] (theta) {$\psi$};
    \edge {gamma} {Zm,Zn};
    \edge {phi} {Xm};
    \edge {theta} {Ym,Yn};    
  }
  \caption{PGM Where Causality Is ``not identifiable''}
  \label{causation_hardpgm}  
\end{figure}

The simplest graphical model where causation becomes
impossible even with unlimited samples is shown in Figure~\ref{causation_hard}.  This fact is demonstrated in the do-calculus
by the fact that there is no way to apply the 3 rules in order to
obtain $\P{y|do(t)}$.

In this problem we consider $T$ to be have  two states
and $Y$ to have two states, but the latent variable or unobserved
confounder $Z$ is of arbitrary complexity.  This reflects many real
life problems e.g. $T$ could represent the presence of some substance
in a person's diet (so it is binary), $Y$ could represent some binary
health outcome and $Z$ could represent socio-economic circumstances of
a person affecting both $T$ and $Y$.

Following the same prescription as before; to find the same solution
using a Bayesian approach we first apply 
\textbf{CausalBayesConstruct} on CGM Figure~\ref{causation_hard} to
produce the PGM in Figure~\ref{causation_hardpgm}. We 
then explicitly parameterize the model and write out the three model
components, the post intervention predictive component, the likelihood component and the
prior component. The first step of parameterization is complicated by
the fact that $Z$ is both latent and high dimensional this results in
posteriors over the parameters that are not-identifiable and high
dimensional, we will also consider a re-parameterization which
partially mitigates these difficulties.

We use the following parameterization:
\eqn{
\P{y|t,z,\bpsi}&=\bpsi_{t,z}^{y}(1-\bpsi_{t,z})^{1-y} =
\P{y^*|t^*,z^*,\bpsi}\nonumber\\
\P{z|\bgamma} &=\bgamma_{z^*} = \P{z^*|\bgamma}, \nonumber\\
\P{t|z,\bphi} &= \bphi_{z}^{t}(1-\bphi_{z})^{1-t}. \nonumber
}
The post intervention predictive is:
\eqn{
\P{&y^*|t^*,\bpsi,\bgamma}  = \sum_{z^*} P(y^*|t^*,z^*,\bpsi)P(z^*|\bgamma) \nonumber\\
& = \sum_{z}  \bpsi_{t^*,z}^{y^*} (1-\bpsi_{t^*,z})^{1-y^*} \bgamma_{z} \nonumber\\
& \equiv  \bPsi_{t^*,y^*}\nonumber
}

We introduce the low dimensional $\bPsi$ as a re-parameterization of
$\bpsi,\bgamma$ as statistically identifying this parameter is sufficient for making
causal inference.

Unfortunately when we write the likelihood we see we cannot identify
this parameter, but rather a different low dimensional function of $\bpsi,\bphi, \bgamma$:
\begin{align*}
&\P{y^*,t^*|\bpsi,\bphi,\bgamma}  \\
&= \sum_{z} \P{y^*|t^*,z,\bpsi}  \P{t^*|z,\bphi}\P{z|\bgamma} \\
& = \sum_{z}  \bpsi_{t^*,z}^{y^*} (1-\bpsi_{t^*,z})^{1-y^*} \bphi_{z}^{t^*}(1-\bphi_{z})^{1-t^*} \bgamma_{z^*}\\
& \equiv \bOmega_{t^*,y^*}.
\end{align*}

We introduce the low dimensional $\bOmega$ as a re-parameterization of
$\bpsi,\bphi, \bgamma$ as this parameter is identifiable.

We can now see the difficulty in this problem, we need $\bPsi$ but can
only infer $\bOmega$.  Both, $\bOmega$ and $\bPsi$ are different low dimensional projections of
$(\bpsi, \bphi, \bgamma)$; $\bOmega$ is identifiable and $\bPsi$ is
causally relevant, they are related due to the fact that they are both
functions of $\bpsi$ and $\bgamma$, so it may be reasonable to specify
a joint prior $\P{\bPsi,\bOmega}$ giving:
\eqn{
\P{&y^*|t^*,\by,\bt} \nonumber \\
&= \int_\Psi \int_{\bOmega} \P{y^*|t^*,\bPsi}\P{\bPsi|\bOmega} \P{\bOmega|\by,\bt} d\bOmega d\bPsi. \nonumber
}

In specifying priors for this problem, we may reasonably use default
priors (e.g. flat priors) for $\bOmega$ (or even take a point estimate)
as it is identifiable with modest 
data sets.  On the other hand $\P{\bPsi|\bOmega}$ will be completely
unaffected by the data so it is an extremely important that any
information concerning on how $\bOmega$ affects knowledge of $\bPsi$ is
carefully assessed; in many instances this will be considered too
difficult to reasonably attempt, e.g. it may be that
$\P{\bPsi|\bOmega}=\P{\bPsi}$ in which case the data adds no value to causal problems at all.

\subsection{The Front Door Rule}

\begin{figure}
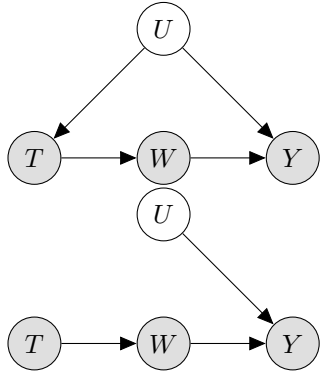

  \centering
  \tikz{ %
    \node[latent] (Um) {$U$} ; %
    \node[obs, below=of Um] (Dm) {$W$} ; %
    \node[obs, left=of Dm] (Xm) {$T$} ; %
    \node[obs, right=of Dm] (Ym) {$Y$} ; %
    \edge {Um} {Xm} ; %
    \edge {Um} {Ym} ; %
    \edge {Xm} {Dm} ; %
    \edge {Dm} {Ym} ; %
  } 
  \tikz{ %
    \node[latent] (Um) {$U$} ; %
    \node[obs, below=of Um] (Dm) {$W$} ; %
    \node[obs, left=of Dm] (Xm) {$T$} ; %
    \node[obs, right=of Dm] (Ym) {$Y$} ; %
    \edge {Um} {Ym} ; %
    \edge {Xm} {Dm} ; %
    \edge {Dm} {Ym} ; %
  }
  \caption{A CGM for the Front Door Rule: Top the original graph, Bottom:
    the mutilated graph}
  \label{cgmfront}  
\end{figure}

Imagine that our observations are generated by the CGM given in
Figure~\ref{cgmfront}, which is the graph that requires the front door
rule.  The front door rule is remarkable in that it shows that a graph
quite similar to Figure~\ref{causation_hard} does allow causation to be
identifiable, the only difference being another observed node between
the treatment and the outcome.

Using the do-calculus gives (detailed steps in \cite{pearl1995causal}):
\eqn{
\P{ y|{\rm do}(T=t) } = \sum_{w} \P{w |t } \sum_{t'} \P{  y|t',w }
  \P{t'}. 
\label{dofd}
}

\begin{figure}[hb!]
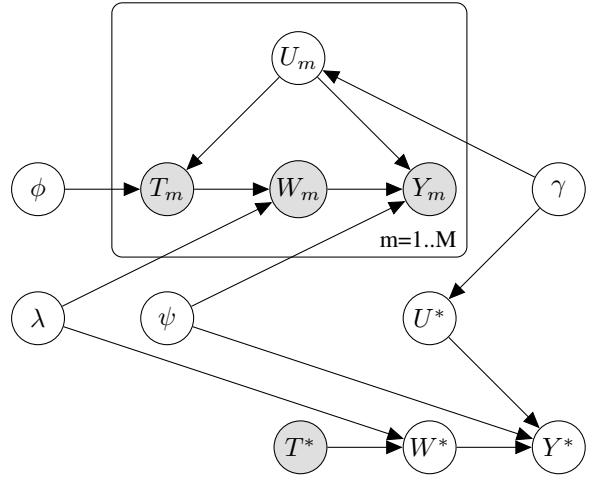

  \centering
  \tikz{ %
    \node[latent] (Um) {$U_m$} ; %
    \node[obs, below=of Um] (Dm) {$W_m$} ; %
    \node[obs, left=of Dm] (Xm) {$T_m$} ; %
    \node[obs, right=of Dm] (Ym) {$Y_m$} ; %
    \edge {Um} {Xm} ; %
    \edge {Um} {Ym} ; %
    \edge {Xm} {Dm} ; %
    \edge {Dm} {Ym} ; %
    \plate[inner sep=0.25cm, xshift=-0.12cm, yshift=0.12cm] {plate1} {(Xm) (Um) (Ym)} {m=1..M};
    \node[latent, below=of Ym] (Un) {$U^*$} ; %
    \node[latent, below=of Un] (Dn) {$W^*$} ; %
    \node[obs, left=of Dn] (Xn) {$T^*$} ; %
    \node[latent, right=of Dn] (Yn) {$Y^*$} ; %
    \edge {Un} {Yn} ; %
    \edge {Xn} {Dn} ; %
    \edge {Dn} {Yn} ; %
%    \plate[inner sep=0.25cm, xshift=-0.12cm, yshift=0.12cm] {plate1} {(Xn) (Un) (Yn)} {n=M+1..N};
    \node[latent, right=of Ym] (gamma) {$\gamma$};
    \node[latent, left=of Xm] (phi) {$\phi$};
    \node[latent, below=of Xm] (theta) {$\psi$};
    \node[latent, left=of theta] (lambda) {$\lambda$}; 
    \edge {gamma} {Um,Un};
    \edge {phi} {Xm};
    \edge {lambda} {Dm,Dn};
    \edge {theta} {Ym,Yn};    
  }
  \caption{A Bayesian Model for the Front Door Rule}
  \label{two_plates_front_door}  
\end{figure}

To find the same solution using a Bayesian approach we first
apply \textbf{CausalBayesConstruct} on CGM Figure~\ref{cgmfront} to
produce the PGM in Figure \ref{two_plates_front_door}.  We then explicitly parameterize the
model and  write out the three model components.   Again we are hampered by the
presence of $u$, but for this problem we can effectively marginalize
and re-parameterize the model to make causation identifiable.  

We use the following parameterization:
\eqn{
&\P{y|w,u,\bpsi}  = \bpsi_{w,u}^{y} (1-\bpsi_{w,u})^{1-y}   =\P{y^*|w^*,u^*,\bpsi}, \nonumber \\
&\P{w|t,\blambda}  =\blambda_{t}^{w}(1-\blambda_{t})^{1-w}\nonumber   =\P{w^*|t^*,\blambda},\\
&\P{u|\bgamma}  = \bgamma_u   =\P{u^*|\bgamma},  \nonumber\\  
&\P{t|u,\bphi}  = \bphi_u^t(1-\bphi_u)^{1-t}.\nonumber
}
The post intervention predictive is:
\eq{
\P{&y^*|t^*,\bpsi,\bgamma,\blambda}  \nonumber\\  
&= \sum_{w^*} P(w^*|t^*,\blambda)\sum_{u^*}\P{u^*|\bgamma}\P{y^*|w^*,u^*,\bpsi}.
}

And the likelihood component is:
\eqn{
&\P{\by,\bt,\bw|\bpsi,\blambda,\bgamma,\bphi}\nonumber \\
= &\prod_m \sum_{u_m} \bpsi_{w_m,u_m}^{y_m}
  (1-\bpsi_{w_m,u_m})^{1-y_m}\blambda_{t_m}^{w_m}(1-\blambda)^{1-w_m} \nonumber \\
& \times\bphi^{t_m}(1-\bphi)^{1-t_m}\bgamma_{u_m}. \nonumber 
}

Unfortunately, the fact that $u$ is both latent and large means that the
posterior over $\bpsi,\bgamma,\bphi$ is both non-identifiable and high dimensional (although the marginal posterior over $\blambda$ is identifiable, since it depends only on $W$ and $T$).  A direct attack would require sophisticated Bayesian approximation methods to
capture the complex structure within the posterior and is not within the scope of this paper. Instead, we note that $u$ can be eliminated from the 2nd term in the post-intervention predictive distribution:

\eqn{
\sum_u&\P{u|\bgamma}\P{y| w,u,\bpsi} \nonumber \\
= &\sum_u\sum_t  \P{u,t|\bgamma,\bphi}\P{y|w,u,\bpsi} \nonumber\\
= &\sum_u\sum_t  \left(\frac{\P{u,t|\bgamma,\bphi}\P{w|t,\blambda}\P{y|w,u,\bpsi}}{\P{w|t,\blambda}}\right) \nonumber\\
= &\sum_t  \left(\frac{\sum_u
    \P{u,t,w,y|\bgamma,\bphi,\blambda,\bpsi}}{\P{w|t,\blambda}}\right)\nonumber \\
= &\sum_t  \frac{\P{t|\bgamma,\bphi}}{\P{w,t|\bgamma,\bphi,\blambda}}\P{y,t,w|\bgamma,\bphi,\blambda,\bpsi} \nonumber \\
= &\sum_t \P{t|\bgamma,\bphi}\P{y|t,w,\bgamma,\bphi,\blambda,\bpsi} \nonumber \\
\ne & \sum_t \P{y,t|w,\bgamma,\bphi,\blambda,\bpsi} ~ {\rm as} ~\P{t|\bgamma,\bphi}\ne \P{t|w,\bgamma,\bphi}\nonumber 
}

We explicitly note this is not a joint distribution of $y,t|w$. Although it is a purely probabilistic expression, it is a
distinctly \emph{Pearlian} one.  We re-parameterize:
\eqn{
& \bOmega_{y,t,w}  \equiv \nonumber \\
& \frac{\sum_u \bpsi_{w,u}^y(1-\bpsi_{w,u})^{1-y}
  \blambda_t^w(1-\blambda_t)^{1-w} \bphi_u^t(1-\bphi_u)^{1-t} \bgamma_u }{\blambda_{t}^{w}(1-\blambda_{t})^{1-w}}\nonumber 
}
This allows us to write the post-intervention predictive distribution as,
\eq{
\P{&y^*|t^*,\bOmega,\blambda}  = \sum_{w} \blambda_{t^*}^{w}(1-\blambda_{t^*})^{1-w}\sum_t \bOmega_{y,t,,w}\nonumber \\
=& \sum_u \sum_w  \bpsi_{w,u}^y(1-\bpsi_{w,u})^{1-y}  \blambda_t^w(1-\blambda_t)^{1-w} \bgamma_u
}
and the likelihood becomes,
\eq{
\P{\by,\bt,\bw|\bOmega,\blambda} =\prod_m \bOmega_{y_m,t_m,w_m}\blambda_{t_m}^{w_m}(1-\blambda_{t_m})^{1-w_m}.
}
As we have an expression for the causal
quantities in terms of $\bOmega,\blambda$ we are able to
identify the parameters needed to make causal estimates.

Finally we establish that the Bayesian solution converges to
the unusual do-calculus expression given in Equation~\ref{dofd}.  The expression
is unusual due to the fact that $t$ is both conditioned on and marginalized.

Again by de Finetti's strong law of large numbers as
as $M\rightarrow \infty$ the the posterior concentrate
on a single point $\hat{\bOmega},\hat{\blambda}$, consequently:

\eqn{
{\rm lim}_{M\rightarrow \infty} \sum_{t'}\hat{\bOmega}_{y,t',w}\rightarrow \sum_{t'}\P{t'}\P{  y|t',w } \nonumber 
}
and:
\eqn{
{\rm lim}_{M\rightarrow \infty} \hat{\blambda}_{t}^{w}(1-\hat{\blambda}_{t})^{1-w} \rightarrow \P{w |t} \nonumber 
}

We once again obtain agreement with the Bayesian solution and the do-calculus:
\eqn{
&{\rm lim}_{M\rightarrow \infty}  \sum_{w} \hat{\blambda}_{t^*}^{w}(1-\hat{\blambda}_{t^*})^{1-w}\sum_t \hat{\bOmega}_{y,t,w} \nonumber \\
&\rightarrow 
\sum_w \P{w|t}\sum_{t'}\P{t'}\P{  y|t',w } \nonumber 
}

Again due to the rapid convergence there will be good agreement also
for small samples.

\section{Conclusion}
The paper shows that it is possible to arrive at the same solution for
causal problems using both the do-calculus and Bayesian theory, the
key insight required for the Bayesian formulation is that the probabilistic
graphical model must model both the pre-intervention and post
intervention world separately, this is perhaps the major contribution
we make otherwise our conclusion is similar to
\cite{lindley1981role}.  Even though it has been long suggested 
that probability alone was insufficient for modeling causal systems
(e.g. \cite{pearl1995causal,pearl2001bayesianism,pearl2009,pearl2014comment} ) it is perhaps unsurprising that Bayesian theory can
accommodate these situations given its long history as arising from axiomizing
reasoning under uncertainty (see \cite{de2017theory,de1980foresight,lindley2000philosophy,bernardo2009bayesian}).  

There remains work to be done for a full unification.  The key benefits
of the Bayesian approach are correct finite sample behavior, and the
ability to produce inference under situations which would be deemed
unidentifiable under the do-calculus; this may be possible because
Bayesian approaches are able to assume that conditional
independence assumptions hold with high probability or approximately
hold and then proceed, where the do-calculus applies hard tests
of conditional independence.

The do-calculus has advantages 
in its ability to discard variables and reduce dimensionality and in
its ability to identify re-parameterizations such as in the case of
the front door rule which enable causality to be identifiable in
surprising cases. Furthermore, the do-calculus is complete: we can
obtain a point estimate for a causal query, without parametric
assumptions on the model, if and only if we can express the query as a
function of the observable, joint distribution pre-intervention via
repeated application of the do-calculus and standard probability
algebra \cite{huang2006pearl,shpitser2006identification}. There is an
algorithm \cite{shpitser2012identification} that, for a given CGM and
causal query, will determine if the query is identifiable via the
do-calculus and; if it is, returns an expression for it in terms of
the observable pre-interventional distribution.   We speculate the
algorithm could be adapted to the Bayesian context in order to
automate the discovery of parameter transforms like the one we
demonstrated for the front door rule.

A rich program of research is therefore open, some topics we think
deserving of attention are: a 
comparison of the do-calculus and Bayesian approaches for instrumental
variable models, CGM graphs where the 
do-calculus produces multiple solutions, CGM graphs where the do-calculus
cannot identify but can produce bounds, CGM graphs where parametric
assumptions are used in combination with the CGM in order to achieve
identifiability and causal exploration where data is used to discover
links before attempting a causal analysis.  We hope other researchers
will take up these challenges.

\bibliography{literature}

\begin{thebibliography}{}

\bibitem[Bernardo and Smith, 2009]{bernardo2009bayesian}
Bernardo, J.~M. and Smith, A.~F. (2009).
\newblock {\em Bayesian theory}, volume 405.
\newblock John Wiley \& Sons.

\bibitem[Dawid, 2015]{dawid2015statistical}
Dawid, A.~P. (2015).
\newblock Statistical causality from a decision-theoretic perspective.
\newblock {\em Annual Review of Statistics and Its Application}, 2:273--303.

\bibitem[De~Finetti, 1974]{de2017theory}
De~Finetti, B. (1974).
\newblock {\em Theory of probability: A critical introductory treatment},
  volume~6.
\newblock John Wiley \& Sons.

\bibitem[De~Finetti, 1980]{de1980foresight}
De~Finetti, B. (1980).
\newblock Foresight: Its logical laws, its subjective sources (1937).
\newblock {\em Studies in subjective probability}, pages 55--118.

\bibitem[Huang and Valtorta, 2006]{huang2006pearl}
Huang, Y. and Valtorta, M. (2006).
\newblock Pearl's calculus of intervention is complete.
\newblock In {\em Proceedings of the Twenty-Second Conference on Uncertainty in
  Artificial Intelligence}, pages 217--224. AUAI Press.

\bibitem[Jordan, 2004]{jordan2004graphical}
Jordan, M.~I. (2004).
\newblock Graphical models.
\newblock {\em Statistical Science}, 19(1):140--155.

\bibitem[Lindley, 2000]{lindley2000philosophy}
Lindley, D.~V. (2000).
\newblock The philosophy of statistics.
\newblock {\em Journal of the Royal Statistical Society: Series D (The
  Statistician)}, 49(3):293--337.

\bibitem[Lindley et~al., 1981]{lindley1981role}
Lindley, D.~V., Novick, M.~R., et~al. (1981).
\newblock The role of exchangeability in inference.
\newblock {\em The Annals of Statistics}, 9(1):45--58.

\bibitem[Pearl, 1995]{pearl1995causal}
Pearl, J. (1995).
\newblock Causal diagrams for empirical research.
\newblock {\em Biometrika}, 82(4):669--688.

\bibitem[Pearl, 2001]{pearl2001bayesianism}
Pearl, J. (2001).
\newblock Bayesianism and causality, or, why i am only a half-{B}ayesian.
\newblock In {\em Foundations of {B}ayesianism}, pages 19--36. Springer.

\bibitem[Pearl, 2009]{pearl2009}
Pearl, J. (2009).
\newblock {\em Causality: Models, Reasoning, and Inference}.
\newblock Cambridge University Press, New York.

\bibitem[Pearl, 2014]{pearl2014comment}
Pearl, J. (2014).
\newblock Comment: understanding {S}impson’s paradox.
\newblock {\em The American Statistician}, 68(1):8--13.

\bibitem[Peters et~al., 2017]{peters2017elements}
Peters, J., Janzing, D., and Sch{\"o}lkopf, B. (2017).
\newblock {\em Elements of causal inference: foundations and learning
  algorithms}.
\newblock MIT press.

\bibitem[Shpitser and Pearl, 2006]{shpitser2006identification}
Shpitser, I. and Pearl, J. (2006).
\newblock Identification of joint interventional distributions in recursive
  semi-markovian causal models.
\newblock In {\em Proceedings of the National Conference on Artificial
  Intelligence}, volume~21, page 1219. Menlo Park, CA; Cambridge, MA; London;
  AAAI Press; MIT Press; 1999.

\bibitem[Shpitser and Pearl, 2012]{shpitser2012identification}
Shpitser, I. and Pearl, J. (2012).
\newblock Identification of conditional interventional distributions.
\newblock In {\em Proceedings of the twenty-second conference on uncertainty in
  artificial intelligence}, pages 437--444. AUAI Press.

\end{thebibliography}

\end{document}